\pdfoutput=1

\documentclass[11pt]{article}

\usepackage[final]{coling}
\usepackage{times}
\usepackage{latexsym}
\usepackage[T1]{fontenc}
\usepackage[utf8]{inputenc}
\usepackage{microtype}
\usepackage{inconsolata}
\usepackage{graphicx}
\usepackage{booktabs}
\usepackage{CJKutf8}
\usepackage[whole]{bxcjkjatype}
\usepackage{tabularx}
\usepackage[subrefformat=parens]{subcaption}
\usepackage{amssymb}

\title{User Willingness-aware Sales Talk Dataset}

\author{
 \textbf{Asahi Hentona\textsuperscript{1}},
 \textbf{Jun Baba\textsuperscript{1}},
 \textbf{Shiki Sato\textsuperscript{1}},
 \textbf{Reina Akama\textsuperscript{2}}
\\
 \textsuperscript{1}CyberAgent,
 \textsuperscript{2}Tohoku University
\\
\texttt{\{hentona\_asahi, baba\_jun, sato\_shiki\}@cyberagent.co.jp}\\ 
\texttt{akama@tohoku.ac.jp}
}

\begin{document}
\maketitle
\begin{abstract}
User willingness is a crucial element in the sales talk process that affects the achievement of the salesperson's or sales system's objectives. Despite the importance of user willingness, to the best of our knowledge, no previous study has addressed the development of automated sales talk dialogue systems that explicitly consider user willingness. A major barrier is the lack of sales talk datasets with reliable user willingness data. Thus, in this study, we developed a user willingness–aware sales talk collection by leveraging the ecological validity concept, which is discussed in the field of human–computer interaction. Our approach focused on three types of user willingness essential in real sales interactions. We created a dialogue environment that closely resembles real-world scenarios to elicit natural user willingness, with participants evaluating their willingness at the utterance level from multiple perspectives. We analyzed the collected data to gain insights into practical user willingness–aware sales talk strategies. In addition, as a practical application of the constructed dataset, we developed and evaluated a sales dialogue system aimed at enhancing the user's intent to purchase.
\end{abstract}

\section{Introduction}

\begin{figure*}[ht]
  \centering
  \includegraphics[width=0.9\linewidth]{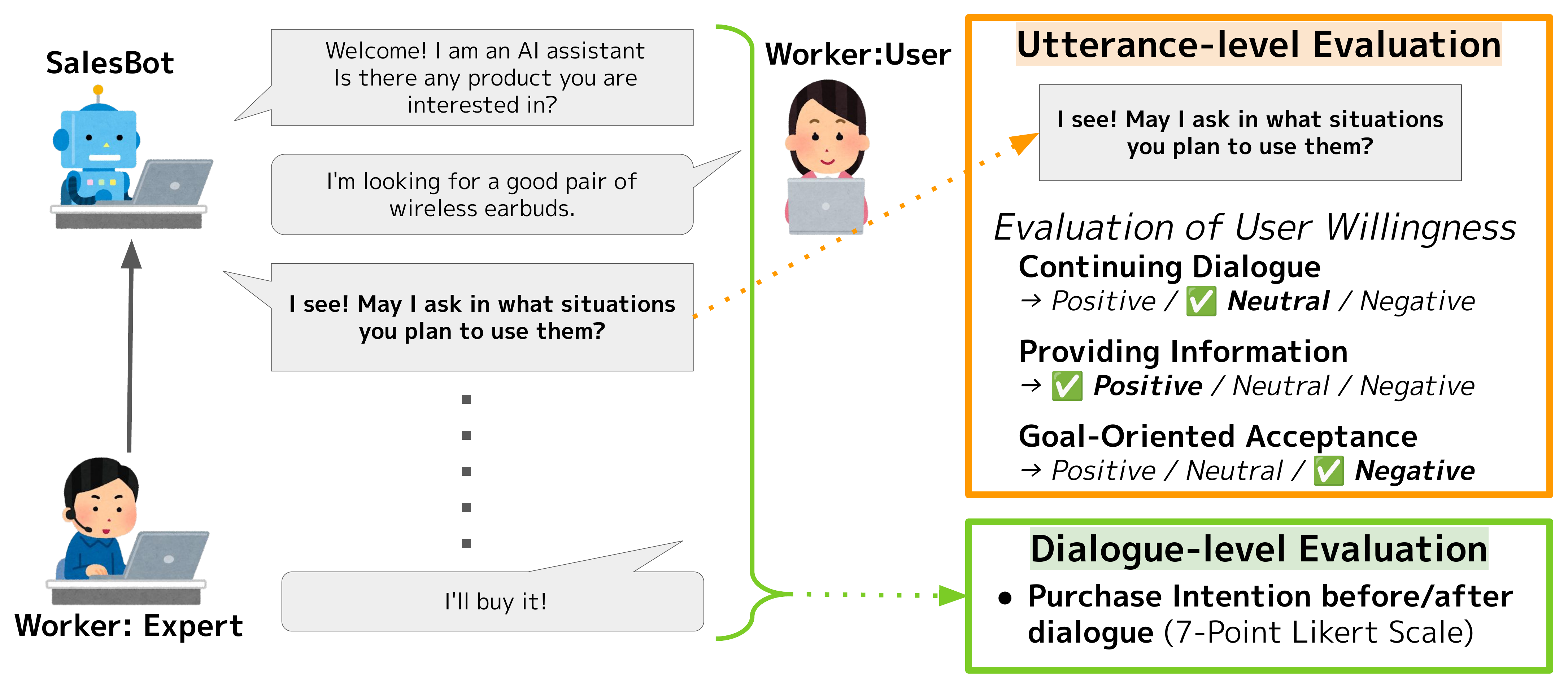}
  \caption{Overview of collecting sales talk dialogue data.}
  \label{fig:top-fig}
\end{figure*}

Sales talk, which involves a series of steps to realize a sale, requires structured communication between a salesperson and a customer~\cite{dubinsky_1981, Hite_1985, dwyer_2013, jung_2022}.
Recently, automated sales talk has attracted increasing attention due to its potential to mitigate labor shortages in developed countries and enhance user experience through constant availability~\cite{hiraoka2016construction, HIRAOKA201683, TIWARI2023118775, murakhovska-etal-2023-salespeople}.

In sales talk, it is important to recognize the various types of user willingness, which significantly affects how salespersons (or dialogue systems) can achieve their objectives. 
We consider at least the following three types of user willingness to be significant: (1) willingness to engage in a dialogue, (2) willingness to provide information, and (3) willingness to accept a salesperson's objectives. 
The willingness to participate in a dialogue, also referred to as user engagement, has been investigated extensively in open-domain dialogue systems~\cite{yu-etal-2016-wizard, zhang-etal-2018-personalizing, see-etal-2019-makes, Ghazarian_Weischedel_Galstyan_Peng_2020}.
A lack of this trait can terminate the dialogue prematurely. 
The willingness to provide information is related to the user's tendency to communicate their needs and desires to the sales talk dialogue system, which allows the system to understand the user's requirements and highlight product benefits effectively, thereby enhancing both persuasiveness and user satisfaction~\cite{wang-etal-2019-persuasion, SUN2022121596, Berkovsky2012}. 
The ultimate objective of sales talk is to motivate a positive decision regarding the purchase of a product~\cite{jung_2022}. 
The willingness to accept a salesperson's objectives involves the user's readiness to embrace the system's primary objective, i.e., making a purchase.

As discussed previously, user willingness affects sales significantly; 
however, to the best of our knowledge, the development of automated sales talk dialogue systems that explicitly involve user willingness has not been investigated to date due to the scarcity of sales talk datasets that contain reliable user willingness data. 
To accurately collect subjective data, including user willingness, from sales talk participants, it is important to consider \emph{ecological validity}~\cite{brunswik_ev_1940, brunswik_ev_1952}. 
The ecological validity concept, which refers to the applicability of experimental results to real-world users that interact with the system in their daily lives, has primarily been investigated in the human–computer interaction (HCI) field. 
Its importance is particularly pronounced when the target task is closely related to practical applications, e.g., sales talks (Section~\ref{subsec:ecological}).

Therefore, in this study, we constructed a user willingness-aware sales talk dataset to advance research on practical sales talk dialogue systems with high ecological validity guided by HCI principles. 
Our approach emphasizes the abovementioned three types of user willingness, which are essential in real-world sales interactions, and creates a dialogue environment that closely simulates actual dialogue scenarios to elicit natural user willingness, where participants evaluate willingness at the utterance level from multiple perspectives. 
To the best of our knowledge, this is the first effort to construct a sales talks dataset annotated with user willingness data.\footnote{\url{https://github.com/CyberAgentAILab/salestalk-dataset}}
The proposed dataset was analyzed to derive valuable insights into practical strategies for user willingness–aware sales talk. 
In addition, as a practical application of the proposed dataset, we developed a sales dialogue system to enhance user purchase intentions. 
The experimental results demonstrate that a dialogue model based on the GPT-3.5 large language model integrates utterance-level user willingness labels and insights from the proposed dataset into its dialogue strategy and improves purchase intentions.

\section{Related Work\label{sec:related-works}}
\subsection{Sales Dialogue Dataset\label{subsec:prior-dataset}}
Several studies involving salespersons in dialogue experiments and collecting human-to-human sales talk dialogues have been conducted previously. 
For example, \citet{hiraoka2016construction} constructed a persuasive dialogue dataset for camera sales. 
They collected persuasive dialogues conducted between customers interested in purchasing a camera and salespersons intending to sell a specific camera. 
In addition, \citet{TIWARI2023118775} constructed a large personalized persuasive dialogue dataset for mobile device purchases, and they presented a persuasive dialogue task for buying a similar product when no products satisfying the customer's specifications were available. However, these two datasets do not include user willingness data, which is crucial in terms of achieving the objectives of persuaders and salespersons. In addition, these datasets comprise human-to-human dialogues, which differ in characteristics from human-to-system dialogues~\cite{serban_survey_2018} which are the focus of the current study. This study focuses on bridging these gaps and collecting data that more closely resembles the real-world environments in which dialogue systems are utilized.

\subsection{Ecological Validity\label{subsec:ecological}}
Ecological validity, as defined by~\citet{brunswik_ev_1940, brunswik_ev_1952}, refers to the statistical correlation between a proximal cue and the distal variable it relates to. In other words, ecological validity measures the extent to which experimental results are applicable to real-world scenarios, where the users interact with systems in their daily lives. This concept is important in both psychology and HCI research because it ensures that research findings can be utilized effectively in practical applications.

Laboratory experiments, although controlled and convenient, frequently fail to replicate the complexities and nuances of real-world interactions~\cite{walf_ev_1989}. 
For example, \citet{levitt_ecological_2007} found that laboratory experiments can exaggerate the significance of prosocial behavior compared with real-world interactions. Such discrepancies address the issue of generalizing lab results to practical settings. In addition, significant progress has been made in natural language processing and dialogue systems; however, there is a clear opportunity to enhance ecological validity further.

In many previous studies, data and user behaviors were collected and evaluated in controlled settings. 
However, despite their usefulness, such studies may not fully capture the complexities of real-world interactions. 
\citet{vries_lui_2020} discussed the issue of low ecological validity of benchmarks and the factors that contribute to low ecological validity. 
They stated that many current researchers employ artificial tasks and synthetic language, do not work with prospective users, prepare participants using scripts, and employ single-turn interfaces, which diverge from real-world language user interface cases. 
As a result, the practical applicability of research findings is limited. In addition, they proposed an ideal methodology to enhance ecological validity to narrow the gap between controlled experiments and real-world applications. 
Their approach involves identifying the target user group and tasks, collecting data using Wizard of Oz simulations~\cite{dahlback_woz_1993, frommherz_ev_2021}, training models based on the collected data, and evaluating user satisfaction comprehensively. 
This method attempts to create realistic and applicable benchmarks to evaluate dialogue systems. 
Inspired by these previous studies, we designed a combined data-collection and user-study process to develop a sales dialogue system to maximize ecological validity.

\section{Dataset Design\label{sec:design}}
In this study, we collected sales talks data that included user willingness information to drive research on practical and user willingness–aware sales talk dialogue systems. As discussed previously, sales talks data must be collected in an environment that is sufficiently similar to real-world dialogue situations to elicit natural user willingness data. In the following, we first outline the envisioned real-world situations in which sales talk dialogue systems are deployed, which is followed by a discussion of the dataset construction policies that were designed to effectively mirror the envisioned situations.

\subsection{Envisioned Situation}
We envisioned a situation where a sales dialogue system is installed on the product page of an e-commerce site or on the landing page (LP) of an advertisement, and the system leads the site visitors (i.e., the users) to purchase products that satisfy their requirements through sales talks. Realizing such sales dialogue systems is crucial for the advertising industry because they could be deployed on LPs and e-commerce sites, which have conventionally been unreachable by salespersons. The actors and properties of sales talk in such situations are described in the following.

\paragraph{Participants.}
Each dialogue occurs between a sales talk dialogue system and a human user.

\paragraph{Beginning and end.}
The user may be unclear about what they wish to ask; thus, it is natural to begin a dialogue by having the system actively initiate an utterance. Note that the user can leave the dialogue at any time, and the dialogue is terminated when the user leaves or purchases a product.

\paragraph{Target users.}
We must acknowledge that user are not necessarily motivated to engage in dialogues. As a result, some users may leave dialogues quickly or may be unwilling to disclose their information. Similarly, users are not necessarily highly motivated to purchase; thus, some users may not buy products.

\paragraph{User willingness.}
User willingness is complex and multifaceted, and it can fluctuate as the dialogue proceeds. For example, some users strongly desire to participate in a dialogue even if they do not intend to purchase a product. In addition, users who are initially reluctant to participate in a dialogue may become interested in purchasing a product as the dialogue progresses.

\subsection{Construction Policies}
\paragraph{Participants.}
We determined that our dataset construction process should utilize the Wizard of Oz method~\cite{Kelley_1984_woz} to collect near-realistic user-side data in sales talks between human users and sales talk systems while ensuring that the sales-side utterances are of sufficient quality.

\paragraph{Beginning and end.}
The data collection process should be performed with explicit instructions to begin a dialogue from the sales-side and to allow the user-side participants to leave a dialogue at any time.

\paragraph{Target users.}
No restrictions should be imposed on the user's motivation to engage in the dialogue. In particular, the users should not be restricted in terms of whether they ultimately purchase products, unlike the settings of previous studies (Section~\ref{subsec:prior-dataset}).

\paragraph{User willingness.}
We consider that more natural user willingness data can be gathered by conducting dialogues in an environment that satisfies the above three points. Thus, we asked the users themselves to assess the impact of salespersons' utterances on the three types of user willingness, i.e., the willingness to engage in dialogue, the willingness to provide information, and the willingness to accept the salesperson's objective. Given that a user's willingness may fluctuate as the dialogue progresses, it is important to evaluate each individual salesperson's utterance rather than the entire dialogue.

\section{Dataset Construction}
\subsection{Data Collection Procedure\label{subsec:construction-procedures}}
Based on the aforementioned construction policies, we collected the sales talk dialogue data according to the following procedure. Figure~\ref{fig:top-fig} shows an overview of the data collection procedure.

\paragraph{Sales-user pairing:}
To replicate sales situations, the participant in the sales role is paired with the participant in the user role and has a dialogue. Before initiating the dialogue, the user-side assessed their initial purchase intent for the displayed product that the sales-side wants to sell (Section~\ref{subsec:construction-settings}) on a seven-point Likert scale. This was then referenced to measure changes in the user's purchase intent as the sales talk progressed.
 
\paragraph{Dialogue:}
The dialogue was conducted using the Wizard-of-Oz method, which masqueraded the sales-side as the system to the user-side. 
Here, each dialogue began with an utterance from the sales-side. 
In this dialogue, the sales-side attempts to persuade the user-side to purchase the product. 
Note that the user-side can interrupt the dialogue at any time if they feel uncomfortable with the sales talk, and the dialogue is terminated when the user leaves the dialogue or purchases the product.\footnote{This study dealt with a fictional product. Even if the user-role participant decided to purchase the product, no business transaction occurred.}

\paragraph{Evaluating:}
The user-side subjectively evaluated the impact of each sales-side utterance on their willingness. Here, for each of the three types of user willingness, we asked the user to classify the sales-side utterances into the three classes, i.e., positive (had a positive impact), neutral (had no impact at all), and negative (had a negative impact). These evaluations were conducted because utterance-level user evaluation data can be used to develop more effective dialogue systems~\cite{ghazarian-etal-2022-wrong, tsuta-etal-2023-rethinking}. In addition, the user-side assessed their post purchase intent for the product in the same manner.

\subsection{Settings\label{subsec:construction-settings}}
\paragraph{Sales scenario.}
The dialogues were performed in a setting where the user, visiting a webpage featuring information on three different types of fictional wireless earphones, engaged in a text chat with a dialogue system integrated into the webpage. Here, we selected the earphone sales setting for two reasons. First, given the technical nature of wireless earphones, considerable expertise is required to understand their features. Products requiring specialized knowledge tend to render users more susceptible to persuasive dialogue~\cite{jung_2022, Wilson1993}, thereby making them ideal for our study, which focuses on sales communication. Second, wireless earphones appeal to a broad demographic, transcending age, sex, and industry distinctions, which is expected to enhance the generalizability of the study's findings. These three products were priced at 11,000, 22,000, and 33,000 yen, which are considered moderate price points for this category of products.

\paragraph{Sales-side person.}
We recruited five fluent Japanese speakers with sales experience as participants via crowdsourcing.\footnote{\url{https://www.lancers.jp/}.} 
To ensure the quality of their involvement in the sales discourse, we provided two hours of extensive experimental guidance, including guidance on crafting talk scripts, engaging in dialogue practice, and learning about product information and product categories. In addition, prior to performing the data collection process, we shared insights into the criteria from the user's perspective to evaluate the sales-side utterances, instructing the sales-side participants to aim for the highest possible assessments.

\paragraph{User-side person.}
A total of 109 user-side participants fluent in Japanese were recruited using the same crowdsourcing platform.

\begin{table}[t]
\centering
\small
\begin{tabular}{lrrrr}
\toprule
             & \multicolumn{1}{l}{Total} & \multicolumn{1}{l}{Mean} & \multicolumn{1}{l}{Max} & \multicolumn{1}{l}{Min} \\
\midrule
\# dialogues          & 109                     & -                      & -                      & -                      \\
\# success dialogues      & 63                     & -                       & -                       & -                 \\
\# utterances          & 3289                   & 30.2                   & 92                     & 11                     \\
\qquad --- User     & 1144                   & 10.5                   & 41                     & 3                      \\
\qquad --- Sales    & 2145                   & 19.7                   & 52                     & 7                      \\
\# tokens        & 54301                  & 498.2                  & 1406                   & 153                    \\
\bottomrule
\end{tabular}
\caption{Statistics of sales talk dataset. We used MeCab\cite{kudo-etal-2004-applying} with UniDic\cite{den-etal-2008-proper} as the tokenizer.\label{tab:dataset-stat}}
\end{table}

\subsection{Dataset Overview}
Table~\ref{tab:dataset-stat} shows the statistical information of the sales dialogue dataset, and a portion of the sales dialogue data included in the dataset is shown in Table~\ref{tab:dialogue-example}. During the dialogue collection process, five experts played the role of the sales-side, and 109 crowd-workers participated as the users. The dialogue collection task was conducted for approximately one month, and a total of 109 dialogues were collected. Here, success in a dialogue was defined as observing an improvement in the users' purchase intentions through the sales talk, i.e., if there was an increase in the seven-point Likert scale rating after the dialogue compared to before. Based on this definition, 63 dialogues (representing 57.8\% of the total) were successful in terms of enhancing the users' purchase intentions.

\begin{table}[t]
    \small 
    \begin{tabularx}{\linewidth}{@{}rXl@{}}
    \toprule
    \textbf{Speaker} & \multicolumn{1}{c}{\textbf{Utterance}} \\ 
    \midrule
        Sales & 
        \textit{Hello.} 
        \textit{What are you looking for?} 
        \\ \specialrule{0pt}{3pt}{0pt}
        User & 
        \textit{I'm looking for earphones with good sound quality. However, I don't have much budget.} 
        \\ \specialrule{0pt}{3pt}{0pt}
        Sales & 
        \textit{Thank you. I would love to help you find the perfect earphones for you.} 
        \textit{By the way, you said you don't have a budget, how much are you considering?} 
        \\ \specialrule{0pt}{3pt}{0pt}
        User & 
        \textit{I'm considering something around 15,000 to 20,000 yen. However, depending on the performance, I could go up to 25,000 yen.} 
        \\ \specialrule{0pt}{3pt}{0pt}
        Sales &
        \textit{Certainly.} 
        \textit{You said you are looking for earphones with good sound quality, does that mean you are not satisfied with the sound quality of your current earphones?} 
        \\ \specialrule{0pt}{3pt}{0pt}
        User & 
        \textit{Yes, I'm bothered by the noise because the noise canceling performance is weak.} 
        \\ 
        \bottomrule
    \end{tabularx}
    \caption{Sample of the sales talk dialogue dataset. Original data is in Japanese, this is English translation.\label{tab:dialogue-example}}
\end{table}

\section{Dataset Analysis}
The sales dataset's most distinctive feature is its inclusion of data on the three types of user willingness supported by a sophisticated experimental configuration with high ecological validity, i.e., the willingness to engage in dialogue (continuing dialogue (CD) willingness), the willingness to provide information (providing information (PI) willingness), and the willingness to accept the salesperson's goal (goal acceptance (GA) willingness). In this section, we analyze the collected dataset and discuss the factors that contribute to the success of the sales talk, with a specific focus on the acquired user willingness data.

\subsection{Dialogue-level Analysis}
\label{subsec:dialogue-level-analysis}

\begin{figure}[t]
    \centering
    \begin{tabular}{cc}
      \begin{minipage}[t]{0.24\textwidth}
        \centering
        \includegraphics[width=0.8\linewidth]{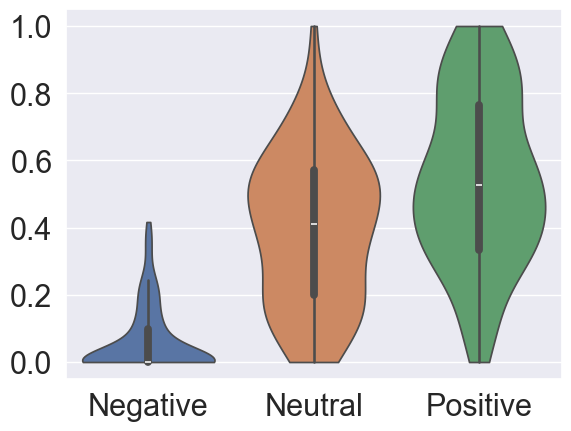}
        \subcaption{CD willingness}
      \end{minipage} &
      \begin{minipage}[t]{0.24\textwidth}
        \centering
        \includegraphics[width=0.8\linewidth]{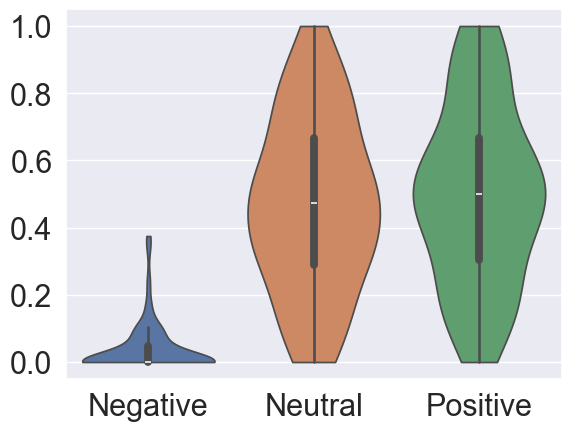}
        \subcaption{PI willingness}
      \end{minipage} \\
      \multicolumn{2}{c}{
      \begin{minipage}[t]{0.24\textwidth}
        \centering
        \includegraphics[width=0.8\linewidth]{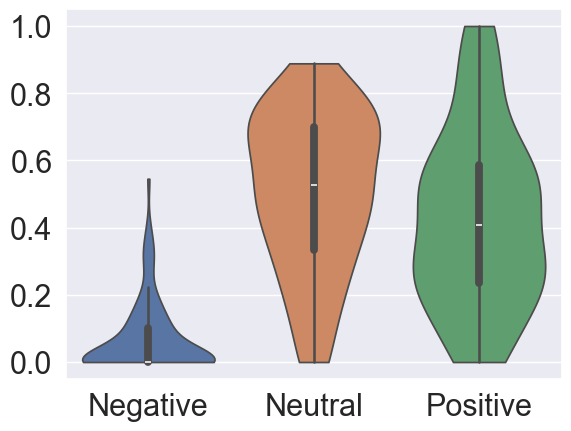}
        \subcaption{GA willingness}
      \end{minipage}}\\
    \end{tabular}
    \caption{Distributions of proportions of user willingness labels (positive, neutral, and negative) across individual dialogues.}
    \label{fig:engagement-violin-plot}
\end{figure}

\paragraph{Distribution of user willingness.}
Figure~\ref{fig:engagement-violin-plot} shows violin plots of the distribution of the proportions of the user willingness evaluation labels (i.e., positive, neutral, and negative) across the individual dialogues. As can be seen, neutral and positive evaluations dominate, with negative evaluations occurring very infrequently, across the three types of user willingness. Note that there are observable differences in the trends of the neutral and positive evaluations between the different types of user willingness. Specifically, CD willingness exhibits a higher tendency to receive positive ratings, with 54.1\% of its ratings falling into this category. In addition, PI willingness, despite receiving many neutral ratings, is rated positively at a level that is comparable to that of CD willingness. Conversely, GA willingness has the lowest proportion of positive evaluations among the three types of user willingness. These findings highlight that, although the negative evaluations are consistently infrequent across all types of user willingness, the distribution of the positive evaluations based varies depending on the specific type of user willingness.

\begin{figure}[t]
  \includegraphics[width=\columnwidth]{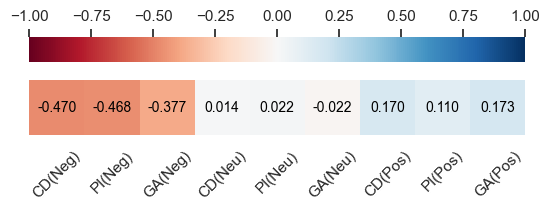}
  \caption{Correlation coefficients between the proportion of user willingness labels and the improvement in the users' purchase intention.}
  \label{fig:engagement_corr}
\end{figure}

\paragraph{Keys to accomplishing sales.}
We conducted a correlation analysis between the proportion of each user willingness evaluation label across the individual dialogues and the improvement in the users' purchase intention, i.e., the amount of change in the seven-point Likert scale rating before and after the dialogue. The results of this correlation analysis are shown in Figure~\ref{fig:engagement_corr}. As can be seen, the positive and neutral evaluations of user willingness exhibit nearly no correlation with improving the users' intent to purchase. Conversely, a negative correlation is observed with the negative evaluations.

\paragraph{Overall.}
These findings suggest that, in terms of enhancing the user's intent to purchase a product, it is crucial to avoid utterances that degrade the user's willingness throughout the dialogue rather than attempting to give utterances that improve or motivate user willingness.

\subsection{Utterance-level Analysis\label{sec:uttr-level-analysis}}
\begin{figure}[t]
    \begin{tabular}{cc}
        \begin{minipage}{0.24\textwidth}
            \includegraphics[width=.9\linewidth]{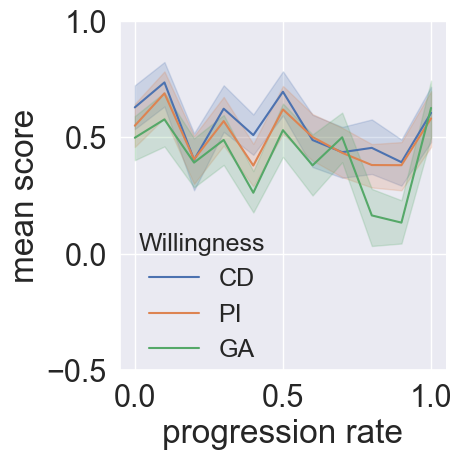}
            \subcaption{Success dialogues}
            \label{fig:engagement_transition_success}
        \end{minipage}%
        \begin{minipage}{0.24\textwidth}
            \includegraphics[width=.9\linewidth]{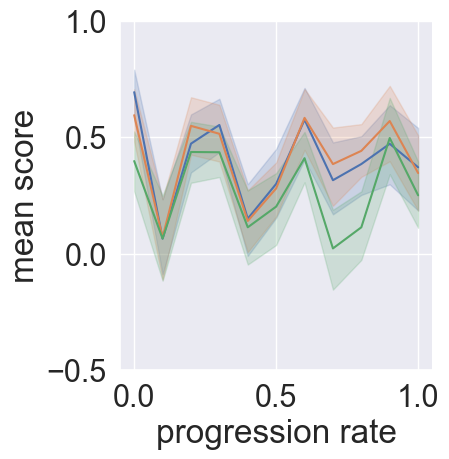}
            \subcaption{Failure dialogues}
            \label{fig:engagement_transition_failure}
        \end{minipage}
    \end{tabular}
    \caption{Average user willingness scores at dialogue progression level.}
\end{figure}

\begin{figure}[t]
    \begin{minipage}{0.24\textwidth}
        \includegraphics[width=0.9\linewidth]{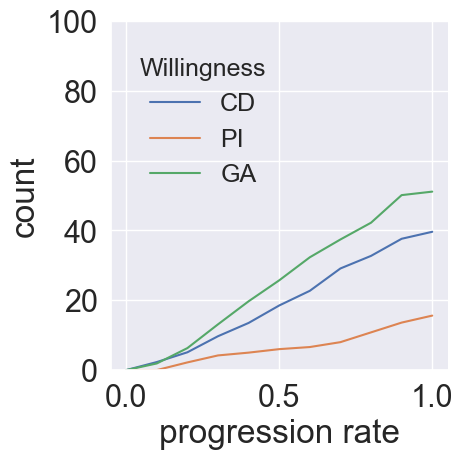}
        \subcaption{Success dialogues\label{fig:engagement_negative_transition_success_large}}
    \end{minipage}%
    \begin{minipage}{0.24\textwidth}
        \includegraphics[width=0.9\linewidth]{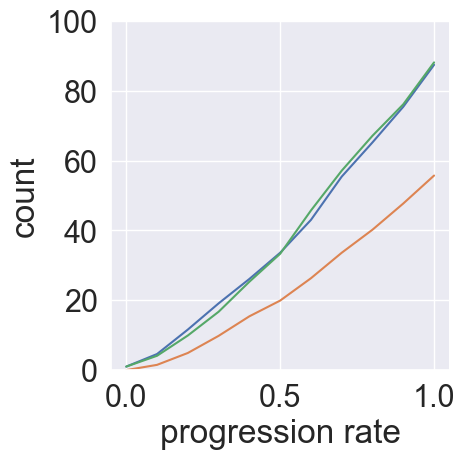}
          \subcaption{Failure dialogues\label{fig:engagement_negative_transition_failure_large}}
    \end{minipage}
    \caption{Cumulative counts of negative user willingness at dialogue progression level.}
\end{figure}

\paragraph{Turn-wise willingness.}
Figure~\ref{fig:engagement_transition_success} shows the average user willingness scores in each stage\footnote{To account for the varying lengths of each dialogue, we normalized them all to a uniform length of $1.0$.}  of all dialogues that enhanced the user's purchase intention, and Figure~\ref{fig:engagement_transition_failure} compares all dialogues that did not enhance the user's intent to purchase. Here, to compute the average willingness scores, the positive labels were assigned a value of $+1$, neutral labels were assigned a value of 0, and negative labels were assigned a value of $-1$. Dialogues that successfully resulted in sales are associated with a noticeable increase in the willingness scores immediately after they begin and just before they end. In particular, the GA willingness scores increased substantially as the dialogues drew to a close. Conversely, in unsuccessful dialogues, all types of willingness scores decreased both shortly after the dialogue began and immediately before it ended. Figures~\ref{fig:engagement_negative_transition_success_large} and~\ref{fig:engagement_negative_transition_failure_large} show the progression of the cumulative counts of the negative willingness assessments over the course of the dialogues, depicting the trends for both successful and unsuccessful dialogues, respectively. In the successful sales dialogues, the total negative assessment counts of the three types of user willingness are approximately one-half to one-third of those observed in the unsuccessful dialogues. Specifically, the negative PI willingness counts are lower during the middle of the dialogues.

\paragraph{Effective sales talk strategies.}
Based on the above analysis, specific strategies for the early, middle, and final stages of the sales dialogue could be effective in terms of increasing the users' intent to purchase.
(1) In the early stages of the dialogue, improving the user's willingness may be effective. When sales talks are initiated by the sales-side, particular consideration should be given to the early stage of the dialogue to ensure that the user does not feel uncomfortable.
(2) In the middle stages of the dialogue, improving and maintaining high PI willingness may be effective. Specifically, probing questions can be utilized to investigate the user's requirements effectively~\cite{jung_2022}.
(3) In the final stages of the dialogue, increasing the user's GA willingness may be effective. For example, it may be useful to conduct closing talks that recommend products that satisfy the user's requirements based on the information gathered in the middle stages of the dialogue.

\section{Experiments}

\begin{table*}[t]
\centering
\small
\begin{tabular}{@{}llcclcc@{}}
\toprule
\multicolumn{1}{c}{\textbf{Model}} & \textbf{Train data}   & \textbf{Willingness} & \textbf{Strategy} & \multicolumn{1}{c}{} & \textbf{Success rate} & \textbf{Avg. \#turns} \\ \cmidrule(r){1-4} \cmidrule(l){6-7} 
GPT-3.5                 & Succcess dialogues    & -                    & -                 &                      & 0.23                  & 9.35                  \\
GPT-3.5W                 & All Dialogues         & \checkmark           & -                 &                      & 0.33                  & 9.23                  \\
GPT-3.5WD      & All Dialogues         & \checkmark           & \checkmark        &                      & 0.44                  & 9.08                  \\ \cmidrule(r){1-4} \cmidrule(l){6-7}
GPT-4o                             & \multicolumn{1}{c}{-} & -                    & -                 &                      & 0.58                  & 5.81                  \\ \bottomrule
\end{tabular}
\caption{Performance of sales dialogue models in user evaluation experiment. \label{tab:user_exp_result}}
\end{table*}

\begin{figure*}[ht]
    \begin{subfigure}[b]{0.24\textwidth}
        \includegraphics[width=\linewidth]{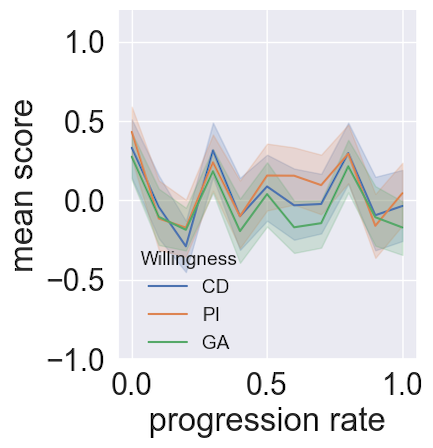}
        \caption{\texttt{GPT-3.5}}
        \label{fig:engagement_transition_success1}
    \end{subfigure}
    \hfill
    \begin{subfigure}[b]{0.24\textwidth}
        \includegraphics[width=\linewidth]{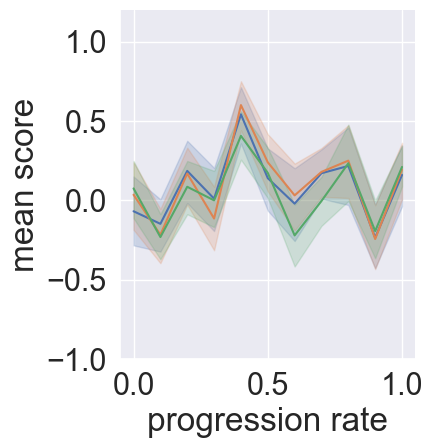}
        \caption{\texttt{GPT-3.5W}}
        \label{fig:engagement_transition_success2}
    \end{subfigure}
    \hfill
    \begin{subfigure}[b]{0.24\textwidth}
        \includegraphics[width=\linewidth]{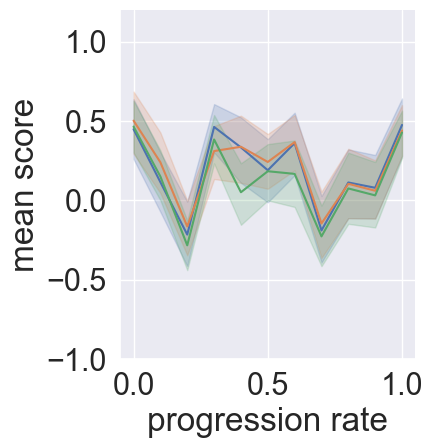}
        \caption{\texttt{GPT-3.5WD}}
        \label{fig:engagement_transition_success3}
    \end{subfigure}
    \hfill
    \begin{subfigure}[b]{0.24\textwidth}
        \includegraphics[width=\linewidth]{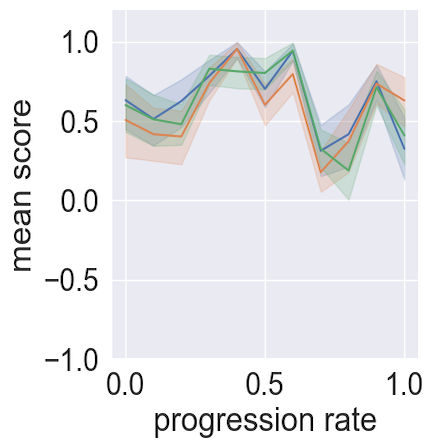}
        \caption{\texttt{GPT-4o}}
        \label{fig:engagement_transition_success4}
    \end{subfigure}
    \caption{Average user willingness for each model in the user evaluation experiment at the dialogue progression level.}
    \label{fig:user_exp_engagement_transitions}
\end{figure*}

This section experimentally demonstrates the dataset's effectiveness in the creation of practical sales dialogue systems. In the following, we highlight the advantages of incorporating near-realistic user willingness data from the constructed dataset into system development, and we describe a user evaluation experiment conducted to evaluate the effectiveness of a sales dialogue system fine-tuned using the dataset developed in this study.

\subsection{Models}
\label{subsec:experiment-models}
In this evaluation, we compared the performance of the following four dialogue systems. Note that the first three systems were developed using OpenAI's GPT-3.5 (\texttt{gpt-3.5-turbo-0125}) fine-tuning API\footnote{\url{https://platform.openai.com}.} with its default settings.

\paragraph{GPT-3.5 as baseline:}
This model is a simple, fine-tuned version of the conventional GPT-3.5 model. Here, we used 63 successful dialogues from the sales dialogue dataset as training data. Note that this model was included in the evaluation to measure the performance of the models fine-tuned. We construct with the aim of measuring performance when only dialogues that ultimately achieved the goal are used as training data, without explicitly considering the utterance level user willingness.

\paragraph{GPT-3.5 with willingness (\texttt{GPT-3.5W}):}
This model was employed to investigate whether considering user willingness at the utterance level can improve the system's ability to achieve its goals. Unlike the baseline model, here, user willingness labels were utilized to fine-tune the GPT-3.5 model. We employed the attribute-conditioned supervised fine-tuning (SFT) method~\cite{dong-etal-2023-steerlm}, which enables a model to generate responses with specified attributes by explicitly providing attribute values for each response in the training data and subsequently training the model on these data. During the model training process, three types of user willingness labels were assigned as attribute values to each response sample in the training dataset. In addition, during the inference process, all responses were generated by specifying attribute values for the three types of willingness to ``positive.'' Note that we used all 109 dialogues in the sales dialogue dataset as the training data.

\paragraph{GPT-3.5 with willingness + dialogue strategy (\texttt{GPT-3.5WD}):} 
The analysis discussed in Section~\ref{sec:uttr-level-analysis} identified the characteristics of effective dialogue strategies in successful sales dialogues, i.e., shifting the emphasis from continuing the dialogue to providing information and finally to goal acceptance. Thus, we investigated the effectiveness of dialogue systems that explicitly incorporate these dialogue strategies, which we could develop with our user willingness–aware dataset. This model was the same as the GPT-3.5W model, except that the attribute values for the attribute-conditioned SFT were changed according to the dialogue progression during inference. Specifically, (1) until the end of the third turn, only the attribute value of the CD willingness was set to positive. (2) From the fourth turn to the sixth turn, only the attribute value of the PI willingness was set to positive. (3) Beginning from the seventh turn, only the attribute value of the GA willingness was set to positive.\footnote{Considered neutral except for the positive label.}

\paragraph{GPT-4o as a reference:}
This model represents the untuned GPT-4o model (\texttt{gpt-4o-2024-05-13}). This untuned model allowed us to assess the proficiency of a cutting-edge largescale language model in executing sales dialogues. Note that the evaluation results obtained with this model are presented only for reference. We believe that the effectiveness of our dataset, specifically its inclusion of the user willingness labels, should be discussed in terms of the evaluation results obtained by the previously described models because GPT-4o has a model scale advantage over GPT-3.5.

\subsection{Evaluation Settings}
We For this experimental evaluation, we recruited 48 participants through crowdsourcing, and each worker engaged in a dialogue with each of the four systems. The workers evaluated the four systems following the same process and settings used for the in the data collection process, as discussed in (Section~\ref{subsec:construction-procedures}). The In addition, the sequence in which the dialogues are were conducted may introduce an order effect bias in evaluating the aforementioned four models . To Thus, to mitigate this potential bias, our the experiment employed a counterbalancing method, whereby the order of the conditions was randomized among the participants.

\section{Experimental Results and Discussion}
Table~\ref{tab:user_exp_result} shows the performance of the four models in the user evaluation experiment in terms of the rate of dialogues in which the user's purchasing intent was increased. In addition, Figure~\ref{fig:user_exp_engagement_transitions} shows the average user willingness scores obtained by the four models at the dialogue progression level.

\subsection{Results on Willingness Label Effectiveness}
\paragraph{Success rate.}
The two models that explicitly considered the utterance-level user willingness labels, i.e., the GPT-3.5W and GPT-3.5WD models, obtained higher success rates than the baseline model. Notably, the GPT-3.5WD model obtained the highest success rate. These results suggest that explicitly considering user willingness can enhance the dialogue success rate effectively. Furthermore, incorporating the dialogue strategies based on our analysis could further increase the effectiveness of utilizing the user willingness labels.

\paragraph{Willingness evaluation.}
Figure~\ref{fig:user_exp_engagement_transitions} shows that the fine-tuned GPT-3.5 model (i.e., the baseline model), which obtained the lowest success rate among the compared models, obtained high user willingness scores in the early stages of the dialogue; however, these scores were low at the end of the dialogue. In contrast, the GPT-3. 5W model, which obtained the second-highest dialogue success rate, obtained lower user willingness scores in the early stages of the dialogue process compared with the GPT-3.5 model; however, these scores were high at the end of the dialogue. Notably, the GPT-3.5WD model, which obtained the highest dialogue success rate among the three models, obtained higher user willingness scores in the early and final stages of the dialogue compared with the other two models. The observed trends for the GPT-3.5WD model correspond with the dialogue strategy intentionally incorporated in its design, as discussed in Section~\ref{subsec:experiment-models}. This alignment indicates that our strategic modifications were implemented successfully in the GPT-3.5WD model. In addition, the GPT-3.5WD model demonstrated superior performance compared with the other models.

\subsection{Results of GPT-4o}
We found that the untuned GPT-4o model obtained the highest dialogue success rate among all four compared models, as shown in Table~\ref{tab:user_exp_result}. In addition, the untuned GPT-4o model exhibited a shorter average turn count than the other models, which suggests that this model conducted dialogues more efficiently to enhance the user's purchase intention; however, minimizing the number of turns in sales dialogues is not the primary focus of this study. Furthermore, Figure~\ref{fig:user_exp_engagement_transitions} also shows that the GPT-4o model obtained higher average user willingness scores throughout the dialogues compared with the other models.

Nevertheless, there may be potential to enhance the sales talk capabilities of the GPT-4o model. For example, a potential area of improvement appears to lie in the relatively low user willingness scores obtained at the end of the dialogues (Figure~\ref{fig:user_exp_engagement_transitions}). Our previous analysis revealed the relationship between the increase in the GA willingness at the end of dialogues and successful outcomes (Section~\ref{sec:uttr-level-analysis}). Thus, optimizing the GPT4o model's ability to increase the users' GA willingness during the final stages of the dialogues could improve the model's success rate considerably. The collected successful sales dialogue data shows a notable increase in GA willingness at the end of the dialogue (Figure~\ref{fig:engagement_transition_success}); thus, training GPT-4o on these data may solve this problem.

\begin{table}[t]
    \small 
    \centering
    \begin{tabularx}{\linewidth}{@{}rXl@{}}
    \toprule
    \textbf{Speaker} & \multicolumn{1}{c}{\textbf{Utterance}} \\ 
    \midrule
        User & 
        \textit{What is the difference between other products?}
        \\ \specialrule{0pt}{3pt}{0pt}
        Sales & 
        \textit{The willingness to continue dialogue is increasing.}
        \\ 
        \bottomrule
    \end{tabularx}
    \caption{Dialogue example with an unclear intention. English translations of the original Japanese data are shown.\label{tab:unclear-dialogue-example}}
\end{table}

\begin{table}[t]
\small
\centering
\begin{tabular}{lrr}
\toprule
\multicolumn{1}{c}{Model} & \multicolumn{1}{c}{\begin{tabular}[c]{@{}c@{}}\# of unclear \\ dialogues\end{tabular}} & \multicolumn{1}{c}{\begin{tabular}[c]{@{}c@{}}\# of successful \\ dialogues\end{tabular}} \\ \midrule
GPT-3.5                    & 7                                                                                              & 1/10                                                                                              \\
GPT-3.5W                   & 6                                                                                              & 4/10                                                                                              \\
GPT-3.5WD                  & 3                                                                                              & 4/10                                                                                              \\
GPT-4o                    & 1                                                                                              & 6/10                                                                                              \\ 
\bottomrule
\end{tabular}
\caption{Number of dialogues containing at least one sales utterance annotated as having unclear intentions.\label{tab:unclear-dialogue-stat}}
\end{table}

\subsection{Analysis of Failure Cases}
We sampled a total of 40 dialogues (10 dialogues for each model) from the dialogue data utilized in the user experiments, and two annotators were recruited to classify the causes of each dialogue failure. As a result, a potential correlation emerged between the number of responses with unclear intentions and the dialogue success rate. An example of a response with an unclear intention is presented in Table~\ref{tab:unclear-dialogue-example}.

In addition, the number of dialogues containing at least one sales utterance annotated as having unclear intentions for each model is shown in Table~\ref{tab:unclear-dialogue-stat}. Here, we deemed utterances to contain unclear intentions if both annotators agreed that the intention was unclear. The results shown in Table 5 suggests that models with lower numbers of these dialogues obtained higher success rates.

\section{Conclusion}
Sales talk is an attractive and practically effective application of dialogue research, and user willingness is a significant factor in practical sales talk. In this study, we developed a sophisticated data collection process based on high ecological validity to faithfully reproduce actual sales dialogue system usage scenarios, and we constructed a sales dialogue dataset that includes subjective user willingness information. The analysis of the constructed dataset yielded various insights into the relationship between user willingness and sales success, as well as valuable insights into effective sales talk strategies. In the user evaluation experiment, we found that explicitly considering user willingness can effectively enhance the dialogue success rate. In addition, the experimental results demonstrated that incorporating sales talk strategies based on our analysis can further increase the effectiveness of considering user willingness in such systems. In the future, we plan to develop a sales dialogue system that can dynamically switch the target user willingness type based on real-time user responses to improve the performance of such sales dialogue systems.

\section*{Limitations}
In this study, we constructed a dataset that assesses user willingness in sales dialogues; however, several limitations must be acknowledged and considered. For example, the dataset was assembled for a specific product category, i.e., wireless earphones, which may limit the generalizability of the results to other product categories. In addition, outcomes may differ based on the price of the product and the characteristics of the target users. In addition, the dataset was collected in the Japanese language, which could limit the applicability and generalizability of the findings to other languages or cultural contexts.  
Finally, the user-side participants in this study did not actually purchase a product; thus, it remains uncertain to what extent their genuine intentions were mirrored in the constructed dataset. Therefore, to address this issue, we plan to conduct a demonstration experiment in which users have the opportunity to actually purchase a product.

\section*{Ethical Considerations}
Prior to conducting the experiments in this study, we obtained ethical approval (CAE-2023-06), and the participants in the dialogue data collection and user evaluation experiments were sourced through a crowdsourcing service. To ensure fair compensation, an hourly wage of 1500 yen was provided, which exceeds the minimum wage in Tokyo, Japan, as of November 2, 2023. This payment was made in accordance with the actual hours worked by the participants. For the dialogue data collection process, we employed the Wizard of Oz method as the dialogue setting. This involved a dialogue between a crowd-worker, who assumed the role of the user, and an experienced salesperson who simulated the sales dialogue system. Upon completion of the experiment, a debriefing was conducted for the user-side participants, where they were informed that their dialogue partner was a human (not a computerized system) to ensure transparency and ethical integrity. In addition, in the user evaluation experiment, we employed largescale language models (i.e., the GPT-3.5 and GPT-4o models). Consequently, there is a concern that harmful statements, inaccurate information, or biases may arise during language generation.

\section*{Acknowledgments}
This work was partly supported by JSPS KAKENHI Grant Numbers JP22K17943.

\bibliography{custom}

\begin{thebibliography}{31}
\providecommand{\natexlab}[1]{#1}

\bibitem[{Berkovsky et~al.(2012)Berkovsky, Freyne, and Oinas-Kukkonen}]{Berkovsky2012}
Shlomo Berkovsky, Jill Freyne, and Harri Oinas-Kukkonen. 2012.
\newblock \href {https://doi.org/10.1145/2209310.2209312} {Influencing individually: Fusing personalization and persuasion}.
\newblock \emph{ACM Trans. Interact. Intell. Syst.}, 2(2).

\bibitem[{Brunswik(1952)}]{brunswik_ev_1952}
E.~Brunswik. 1952.
\newblock \href {https://books.google.co.jp/books?id=flINAQAAMAAJ} {\emph{The Conceptual Framework of Psychology}}.
\newblock International encyclopedia of unified science. University of Chicago Press.

\bibitem[{Brunswik(1940)}]{brunswik_ev_1940}
Egon Brunswik. 1940.
\newblock Thing constancy as measured by correlation coefficients.
\newblock \emph{Psychological Review}, 47(1):69.

\bibitem[{Dahlb{\"a}ck et~al.(1993)Dahlb{\"a}ck, J{\"o}nsson, and Ahrenberg}]{dahlback_woz_1993}
Nils Dahlb{\"a}ck, Arne J{\"o}nsson, and Lars Ahrenberg. 1993.
\newblock Wizard of oz studies: why and how.
\newblock In \emph{Proceedings of the 1st international conference on Intelligent user interfaces}, pages 193--200.

\bibitem[{Den et~al.(2008)Den, Nakamura, Ogiso, and Ogura}]{den-etal-2008-proper}
Yasuharu Den, Junpei Nakamura, Toshinobu Ogiso, and Hideki Ogura. 2008.
\newblock \href {http://www.lrec-conf.org/proceedings/lrec2008/pdf/258_paper.pdf} {A proper approach to {J}apanese morphological analysis: Dictionary, model, and evaluation}.
\newblock In \emph{Proceedings of the Sixth International Conference on Language Resources and Evaluation ({LREC}'08)}, Marrakech, Morocco. European Language Resources Association (ELRA).

\bibitem[{Dong et~al.(2023)Dong, Wang, Sreedhar, Wu, and Kuchaiev}]{dong-etal-2023-steerlm}
Yi~Dong, Zhilin Wang, Makesh Sreedhar, Xianchao Wu, and Oleksii Kuchaiev. 2023.
\newblock \href {https://doi.org/10.18653/v1/2023.findings-emnlp.754} {{S}teer{LM}: Attribute conditioned {SFT} as an (user-steerable) alternative to {RLHF}}.
\newblock In \emph{Findings of the Association for Computational Linguistics: EMNLP 2023}, pages 11275--11288, Singapore. Association for Computational Linguistics.

\bibitem[{Dubinsky(1981)}]{dubinsky_1981}
Alan~J. Dubinsky. 1981.
\newblock \href {https://doi.org/10.1080/08853134.1981.10754192} {A factor analytic study of the personal selling process}.
\newblock \emph{Journal of Personal Selling \& Sales Management}, 1(1):26--33.

\bibitem[{Frommherz and Zarcone(2021)}]{frommherz_ev_2021}
Yannick Frommherz and Alessandra Zarcone. 2021.
\newblock \href {https://doi.org/10.3389/fcomp.2021.686050} {Crowdsourcing ecologically-valid dialogue data for german}.
\newblock \emph{Frontiers in Computer Science}, 3.

\bibitem[{Ghazarian et~al.(2022)Ghazarian, Hedayatnia, Papangelis, Liu, and Hakkani-Tur}]{ghazarian-etal-2022-wrong}
Sarik Ghazarian, Behnam Hedayatnia, Alexandros Papangelis, Yang Liu, and Dilek Hakkani-Tur. 2022.
\newblock \href {https://doi.org/10.18653/v1/2022.findings-acl.331} {What is wrong with you?: Leveraging user sentiment for automatic dialog evaluation}.
\newblock In \emph{Findings of the Association for Computational Linguistics: ACL 2022}, pages 4194--4204, Dublin, Ireland. Association for Computational Linguistics.

\bibitem[{Ghazarian et~al.(2020)Ghazarian, Weischedel, Galstyan, and Peng}]{Ghazarian_Weischedel_Galstyan_Peng_2020}
Sarik Ghazarian, Ralph Weischedel, Aram Galstyan, and Nanyun Peng. 2020.
\newblock \href {https://doi.org/10.1609/aaai.v34i05.6283} {Predictive engagement: An efficient metric for automatic evaluation of open-domain dialogue systems}.
\newblock \emph{Proceedings of the AAAI Conference on Artificial Intelligence}, 34(05):7789--7796.

\bibitem[{Götze et~al.(2022)Götze, Paetzel-Prüsmann, Liermann, Diekmann, and Schlangen}]{gotze_slurk_2022}
Jana Götze, Maike Paetzel-Prüsmann, Wencke Liermann, Tim Diekmann, and David Schlangen. 2022.
\newblock \href {https://aclanthology.org/2022.lrec-1.433} {The slurk {Interaction} {Server} {Framework}: {Better} {Data} for {Better} {Dialog} {Models}}.
\newblock In \emph{Proceedings of the {Thirteenth} {Language} {Resources} and {Evaluation} {Conference}}, pages 4069--4078.

\bibitem[{Hiraoka et~al.(2016{\natexlab{a}})Hiraoka, Neubig, Sakti, Toda, and Nakamura}]{hiraoka2016construction}
Takuya Hiraoka, Graham Neubig, Sakriani Sakti, Tomoki Toda, and Satoshi Nakamura. 2016{\natexlab{a}}.
\newblock Construction and analysis of a persuasive dialogue corpus.
\newblock \emph{Situated Dialog in Speech-Based Human-Computer Interaction}, pages 125--138.

\bibitem[{Hiraoka et~al.(2016{\natexlab{b}})Hiraoka, Neubig, Sakti, Toda, and Nakamura}]{HIRAOKA201683}
Takuya Hiraoka, Graham Neubig, Sakriani Sakti, Tomoki Toda, and Satoshi Nakamura. 2016{\natexlab{b}}.
\newblock \href {https://doi.org/10.1016/j.specom.2016.09.002} {Learning cooperative persuasive dialogue policies using framing}.
\newblock \emph{Speech Communication}, 84:83--96.

\bibitem[{Hite and Bellizzi(1985)}]{Hite_1985}
Robert~E. Hite and Joseph~A. Bellizzi. 1985.
\newblock \href {https://doi.org/10.1080/08853134.1985.10754398} {Differences in the importance of selling techniques between consumer and industrial salespeople}.
\newblock \emph{Journal of Personal Selling \& Sales Management}, 5(2):19--30.

\bibitem[{Jung(2022)}]{jung_2022}
Yeonkwon Jung. 2022.
\newblock \href {https://doi.org/10.1007/978-981-19-0051-8_5} {\emph{Sales Talk}}, pages 87--114.
\newblock Springer Nature Singapore, Singapore.

\bibitem[{Kelley(1984)}]{Kelley_1984_woz}
J.~F. Kelley. 1984.
\newblock \href {https://doi.org/10.1145/357417.357420} {An iterative design methodology for user-friendly natural language office information applications}.
\newblock \emph{ACM Trans. Inf. Syst.}, 2(1):26–41.

\bibitem[{Kudo et~al.(2004)Kudo, Yamamoto, and Matsumoto}]{kudo-etal-2004-applying}
Taku Kudo, Kaoru Yamamoto, and Yuji Matsumoto. 2004.
\newblock \href {https://aclanthology.org/W04-3230} {Applying conditional random fields to {J}apanese morphological analysis}.
\newblock In \emph{Proceedings of the 2004 Conference on Empirical Methods in Natural Language Processing}, pages 230--237, Barcelona, Spain. Association for Computational Linguistics.

\bibitem[{Levitt and List(2007)}]{levitt_ecological_2007}
Steven~D. Levitt and John~A. List. 2007.
\newblock \href {https://doi.org/10.1257/jep.21.2.153} {What do laboratory experiments measuring social preferences reveal about the real world?}
\newblock \emph{Journal of Economic Perspectives}, 21(2):153–174.

\bibitem[{Murakhovs{'}ka et~al.(2023)Murakhovs{'}ka, Laban, Xie, Xiong, and Wu}]{murakhovska-etal-2023-salespeople}
Lidiya Murakhovs{'}ka, Philippe Laban, Tian Xie, Caiming Xiong, and Chien-Sheng Wu. 2023.
\newblock \href {https://doi.org/10.18653/v1/2023.findings-emnlp.657} {Salespeople vs {S}ales{B}ot: Exploring the role of educational value in conversational recommender systems}.
\newblock In \emph{Findings of the Association for Computational Linguistics: EMNLP 2023}, pages 9823--9838, Singapore. Association for Computational Linguistics.

\bibitem[{Sean~Dwyer and Martin(2000)}]{dwyer_2013}
John~Hill Sean~Dwyer and Warren Martin. 2000.
\newblock \href {https://doi.org/10.1080/08853134.2000.10754235} {An empirical investigation of critical success factors in the personal selling process for homogenous goods}.
\newblock \emph{Journal of Personal Selling \& Sales Management}, 20(3):151--159.

\bibitem[{See et~al.(2019)See, Roller, Kiela, and Weston}]{see-etal-2019-makes}
Abigail See, Stephen Roller, Douwe Kiela, and Jason Weston. 2019.
\newblock \href {https://doi.org/10.18653/v1/N19-1170} {What makes a good conversation? how controllable attributes affect human judgments}.
\newblock In \emph{Proceedings of the 2019 Conference of the North {A}merican Chapter of the Association for Computational Linguistics: Human Language Technologies, Volume 1 (Long and Short Papers)}, pages 1702--1723, Minneapolis, Minnesota. Association for Computational Linguistics.

\bibitem[{Serban et~al.(2018)Serban, Lowe, Henderson, Charlin, and Pineau}]{serban_survey_2018}
Iulian~Vlad Serban, Ryan Lowe, Peter Henderson, Laurent Charlin, and Joelle Pineau. 2018.
\newblock \href {https://doi.org/10.5087/dad.2018.101} {A {Survey} of {Available} {Corpora} {For} {Building} {Data}-{Driven} {Dialogue} {Systems}}.
\newblock \emph{Dialogue \& Discourse}, 9(1):1--49.

\bibitem[{Sun et~al.(2022)Sun, Zhang, Zhu, Jiang, and Chen}]{SUN2022121596}
Shiwei Sun, Jin Zhang, Yiwei Zhu, Mian Jiang, and Shuhui Chen. 2022.
\newblock \href {https://doi.org/10.1016/j.techfore.2022.121596} {Exploring users' willingness to disclose personal information in online healthcare communities: The role of satisfaction}.
\newblock \emph{Technological Forecasting and Social Change}, 178:121596.

\bibitem[{Tiwari et~al.(2023)Tiwari, Khandwe, Saha, Ramnani, Maitra, and Sengupta}]{TIWARI2023118775}
Abhisek Tiwari, Abhijeet Khandwe, Sriparna Saha, Roshni Ramnani, Anutosh Maitra, and Shubhashis Sengupta. 2023.
\newblock \href {https://doi.org/10.1016/j.eswa.2022.118775} {Towards personalized persuasive dialogue generation for adversarial task oriented dialogue setting}.
\newblock \emph{Expert Systems with Applications}, 213:118775.

\bibitem[{Tsuta et~al.(2023)Tsuta, Yoshinaga, Sato, and Toyoda}]{tsuta-etal-2023-rethinking}
Yuma Tsuta, Naoki Yoshinaga, Shoetsu Sato, and Masashi Toyoda. 2023.
\newblock \href {https://doi.org/10.18653/v1/2023.ijcnlp-srw.8} {Rethinking response evaluation from interlocutor{'}s eye for open-domain dialogue systems}.
\newblock In \emph{Proceedings of the 13th International Joint Conference on Natural Language Processing and the 3rd Conference of the Asia-Pacific Chapter of the Association for Computational Linguistics: Student Research Workshop}, pages 55--63, Nusa Dua, Bali. Association for Computational Linguistics.

\bibitem[{Vries et~al.(2020)Vries, Bahdanau, and Manning}]{vries_lui_2020}
Harm~De Vries, Dzmitry Bahdanau, and Christopher~D. Manning. 2020.
\newblock \href {https://api.semanticscholar.org/CorpusID:220845833} {Towards ecologically valid research on language user interfaces}.
\newblock \emph{ArXiv}, abs/2007.14435.

\bibitem[{Wang et~al.(2019)Wang, Shi, Kim, Oh, Yang, Zhang, and Yu}]{wang-etal-2019-persuasion}
Xuewei Wang, Weiyan Shi, Richard Kim, Yoojung Oh, Sijia Yang, Jingwen Zhang, and Zhou Yu. 2019.
\newblock \href {https://doi.org/10.18653/v1/P19-1566} {Persuasion for good: Towards a personalized persuasive dialogue system for social good}.
\newblock In \emph{Proceedings of the 57th Annual Meeting of the Association for Computational Linguistics}, pages 5635--5649, Florence, Italy. Association for Computational Linguistics.

\bibitem[{Wilson and Sherrell(1993)}]{Wilson1993}
Elizabeth~J. Wilson and Daniel~L. Sherrell. 1993.
\newblock \href {https://doi.org/10.1007/bf02894421} {Source effects in communication and persuasion research: A meta-analysis of effect size}.
\newblock \emph{Journal of the Academy of Marketing Science}, 21(2):101–112.

\bibitem[{Wolf et~al.(1989)Wolf, Carroll, Landauer, John, and Whiteside}]{walf_ev_1989}
C.~G. Wolf, J.~M. Carroll, T.~K. Landauer, B.~E. John, and J.~Whiteside. 1989.
\newblock \href {https://doi.org/10.1145/67449.67500} {The role of laboratory experiments in hci: help, hindrance, or ho-hum?}
\newblock In \emph{Proceedings of the SIGCHI Conference on Human Factors in Computing Systems}, CHI '89, page 265–268, New York, NY, USA. Association for Computing Machinery.

\bibitem[{Yu et~al.(2016)Yu, Nicolich-Henkin, Black, and Rudnicky}]{yu-etal-2016-wizard}
Zhou Yu, Leah Nicolich-Henkin, Alan~W Black, and Alexander Rudnicky. 2016.
\newblock \href {https://doi.org/10.18653/v1/W16-3608} {A {W}izard-of-{O}z study on a non-task-oriented dialog systems that reacts to user engagement}.
\newblock In \emph{Proceedings of the 17th Annual Meeting of the Special Interest Group on Discourse and Dialogue}, pages 55--63, Los Angeles. Association for Computational Linguistics.

\bibitem[{Zhang et~al.(2018)Zhang, Dinan, Urbanek, Szlam, Kiela, and Weston}]{zhang-etal-2018-personalizing}
Saizheng Zhang, Emily Dinan, Jack Urbanek, Arthur Szlam, Douwe Kiela, and Jason Weston. 2018.
\newblock \href {https://doi.org/10.18653/v1/P18-1205} {Personalizing dialogue agents: {I} have a dog, do you have pets too?}
\newblock In \emph{Proceedings of the 56th Annual Meeting of the Association for Computational Linguistics (Volume 1: Long Papers)}, pages 2204--2213, Melbourne, Australia. Association for Computational Linguistics.

\end{thebibliography}

\newpage
\appendix

\section{User Analysis}
\paragraph{Motivation.}
Users with extremely low prepurchase intent are liable to be largely unaffected by sales talk. Therefore, if the dataset contains many dialogues by such users, measuring the effectiveness of the sales talk can be difficult. This means that the presence of users with low prepurchase intent affects the quality of the dataset.

\begin{figure}[ht]
  \centering
  \includegraphics[width=0.8\columnwidth]{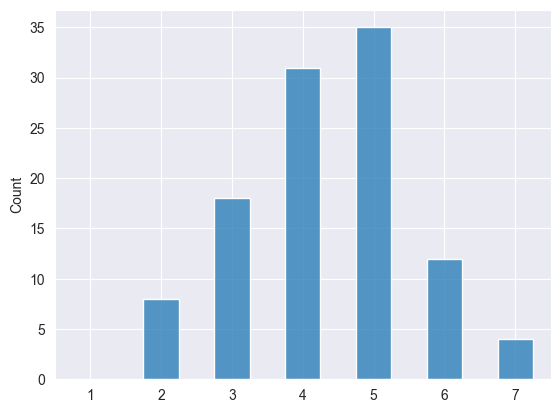}
  \caption{Distribution of user evaluations regarding prepurchase intentions.\label{fig:before_purchase_intention_hist}}
\end{figure}

\paragraph{Result.}
As discussed in Section~\ref{subsec:construction-procedures}, all users participating in the dialogue collection answered a pre-purchase intention to purchase question prior to participating in the sales talk. The distribution of pre-purchase intentions of user-side participants is shown in Figure~\ref{fig:before_purchase_intention_hist}. Only 26 of the 109 users indicated that their pre-purchase intentions were lower than ``4: Neither,'' indicating that the majority of users had moderate or higher pre-purchase intentions. Notably, no user in this dataset gave the lowest score of ``1'' for pre-purchase intention. Based on these findings, we conclude that our dataset is of sufficient quality to allow us to measure the effectiveness of the sales talk.

\newpage
\section{Dialogue Platform}
\begin{figure}[h]
    \centering
    \includegraphics[width=\linewidth]{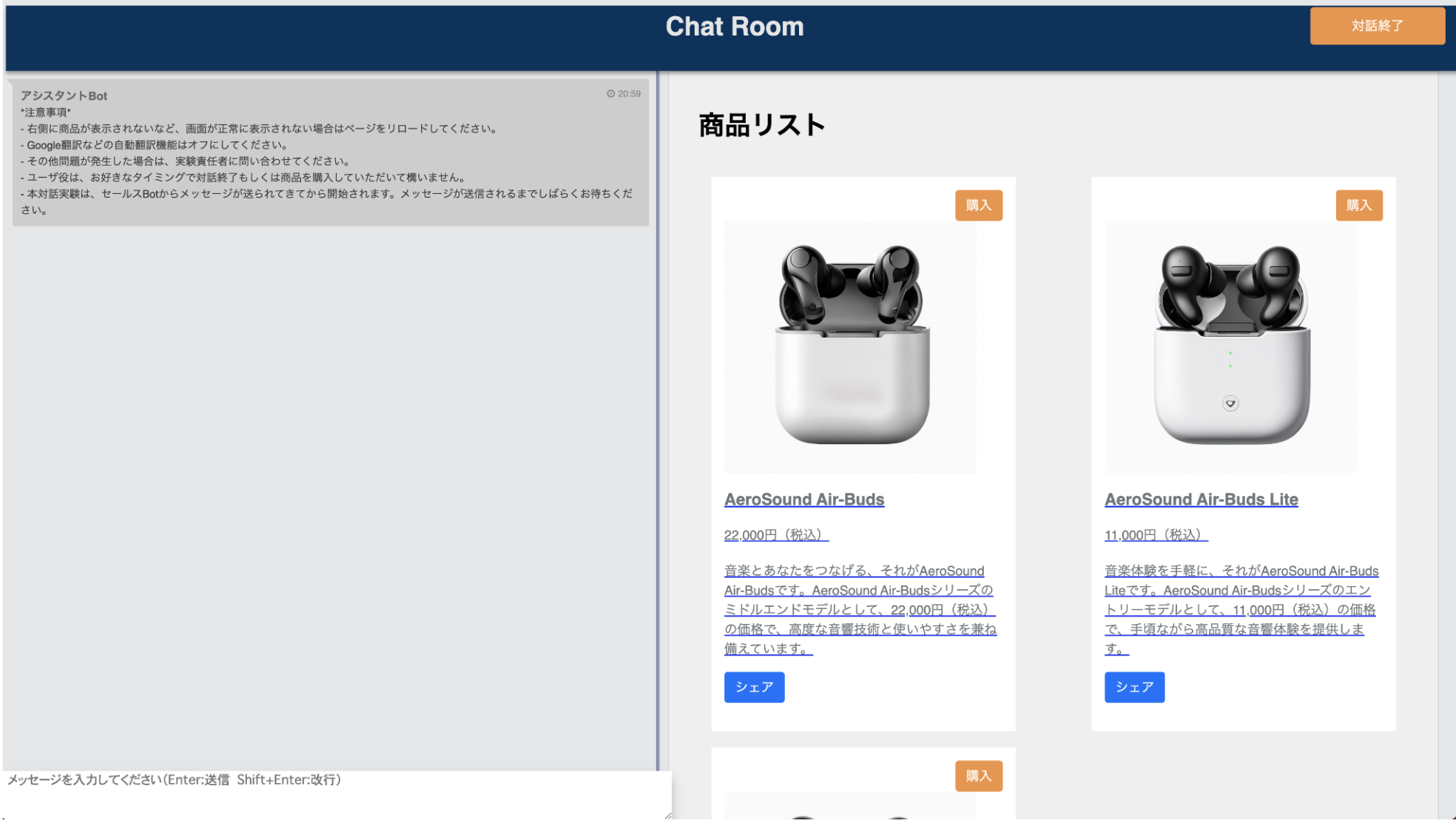}
    \caption{Chat-tool interface for sales talk.}
    \label{fig:chattool}
\end{figure}

The interface of the chat tool used in the experiment is shown in Figure~\ref{fig:chattool}. The text chat tool is a system based on slurk~\cite{gotze_slurk_2022}, and is deployed on a cloud server. In the experiment, when the participants access the chat tool URL, a pair of user role and sales role participants are matched, and the dialogue experiment begins automatically, creating a chat room. In addition to basic text chat features, this chat tool implements functions that are appropriate for our study, such as displaying and sharing product information, and ending the dialogue.

\end{document}